\documentclass[11pt]{article}

\usepackage[final]{acl}

\usepackage{times}
\usepackage{latexsym}

\usepackage[T1]{fontenc}

\usepackage{microtype}

\usepackage{inconsolata}
\usepackage{amsmath}
\usepackage{graphicx}
\usepackage{graphicx}
\usepackage{booktabs}
\usepackage{multirow}
\usepackage[table]{xcolor}
\usepackage[table]{xcolor}
\usepackage{booktabs}
\usepackage{multirow}
\usepackage{array}
\usepackage{graphicx}
\title{RuleTrek: Benchmarking Agentic Rule Reasoning in Complex Scenarios}

\author{
    \textbf{Chenhao Yuan}\textsuperscript{1}\thanks{Equal contribution.},
    \textbf{Bohan Yu}\textsuperscript{1,*},
    \textbf{Yinhao Xu}\textsuperscript{2,*},
    \textbf{Chao Yin}\textsuperscript{2},
    \textbf{Chenxi Zhou}\textsuperscript{1}, \\
    \textbf{Zhaoxuan Wang}\textsuperscript{2},
    \textbf{Shuwen Xu}\textsuperscript{2},
    \textbf{Shaoping Huang}\textsuperscript{1},
    \textbf{Haolin Ren}\textsuperscript{1,2}, \\
    \textbf{Jiaxiang Liu}\textsuperscript{1},
    \textbf{Yao Xu}\textsuperscript{3},
    \textbf{Jun Zhao}\textsuperscript{1,2},
    \textbf{Kang Liu}\textsuperscript{1,2}\thanks{Corresponding author.}
    \\[0.8em]
    \textsuperscript{1}The Key Laboratory of Cognition and Decision
    Intelligence for Complex Systems, \\
    Institute of Automation, Chinese Academy of Sciences,
    Beijing, China \\
    \textsuperscript{2}School of Artificial Intelligence,
    University of Chinese Academy of Sciences, Beijing, China \\
    \textsuperscript{3}Beijing Academy of Artificial Intelligence
    (BAAI), Beijing, China
}

\usepackage{pifont}

\usepackage{amssymb}
\usepackage{xcolor}
\usepackage{booktabs}
\usepackage{tabularx}
\usepackage{array}
\usepackage{amssymb}
\usepackage[table]{xcolor}
\newcommand{\cmark}{%
\textcolor{green!60!black}{\ding{51}}}

\newcommand{\xmark}{%
\textcolor{red!80!black}{\ding{55}}}

\newcommand{\pmark}{\textcolor{orange!90!black}{\ensuremath{\ominus}}}

\usepackage{xcolor}
\usepackage{listings}
\usepackage{tcolorbox}

\newtcolorbox{judgeprompt}[1]{
  title={#1},
  colback=black!2,
  colframe=black!40,
  colbacktitle=black!8,
  coltitle=black,
  fonttitle=\small\bfseries,
  fontupper=\small,
  boxrule=0.4pt,
  arc=1mm,
  left=2.5mm, right=2.5mm, top=2mm, bottom=2mm,
  before upper={\setlength{\parindent}{0pt}\setlength{\parskip}{3pt}}
}

\lstdefinestyle{judgejson}{
  basicstyle=\ttfamily\footnotesize,
  columns=fullflexible,
  keepspaces=true,
  showstringspaces=false,
  upquote=true,
  breaklines=true,
  breakatwhitespace=false,
  aboveskip=3pt,
  belowskip=3pt
}

\newenvironment{judgeitems}{%
  \begin{list}{\textbullet}{%
    \setlength{\leftmargin}{1.2em}%
    \setlength{\labelsep}{0.4em}%
    \setlength{\itemsep}{1pt}%
    \setlength{\topsep}{2pt}%
    \setlength{\parsep}{0pt}%
  }%
}{\end{list}}

\begin{document}
\maketitle
\begin{abstract}

Large language models (LLMs) increasingly
tackle complex real-world tasks in specialized domains.
However, identifying and correctly applying
domain-specific rules remains challenging when
scenario information is distributed across multiple files.
Existing rule-reasoning benchmarks typically
represent scenarios using only a few factual
statements, simplifying the identification
and application of relevant rules.
To address this gap, this paper introduces
\textbf{RuleTrek}, a multi-domain benchmark
for agentic rule reasoning in complex scenarios.
Its rule-grounded construction pipeline
extracts structured rule graphs, generates
and incrementally composes rule-specific
sub-scenarios while ensuring consistency
and solvability.
To systematically evaluate agentic rule
reasoning in complex scenarios, RuleTrek
assesses three capabilities in interactive
workspaces: Rule Identification (RI),
Rule Deduction (RD), and End-to-End
Rule Reasoning (E2E).
Experiments on 11 frontier LLMs show that
the highest RI and RD scores are only
62.47\% and 60.43\%, respectively,
while E2E scores range from 26.97\% to 55.29\%.
These results demonstrate that current
LLM agents face substantial challenges
in rule reasoning over complex scenarios. We make our code and dataset publicly available at
\href{https://github.com/yuan247/RuleTrek}{RuleTrek}.
\end{abstract}

\section{Introduction}
\label{sec:introduction}
\begin{figure}[t]
    \centering
    \includegraphics[width=\linewidth]{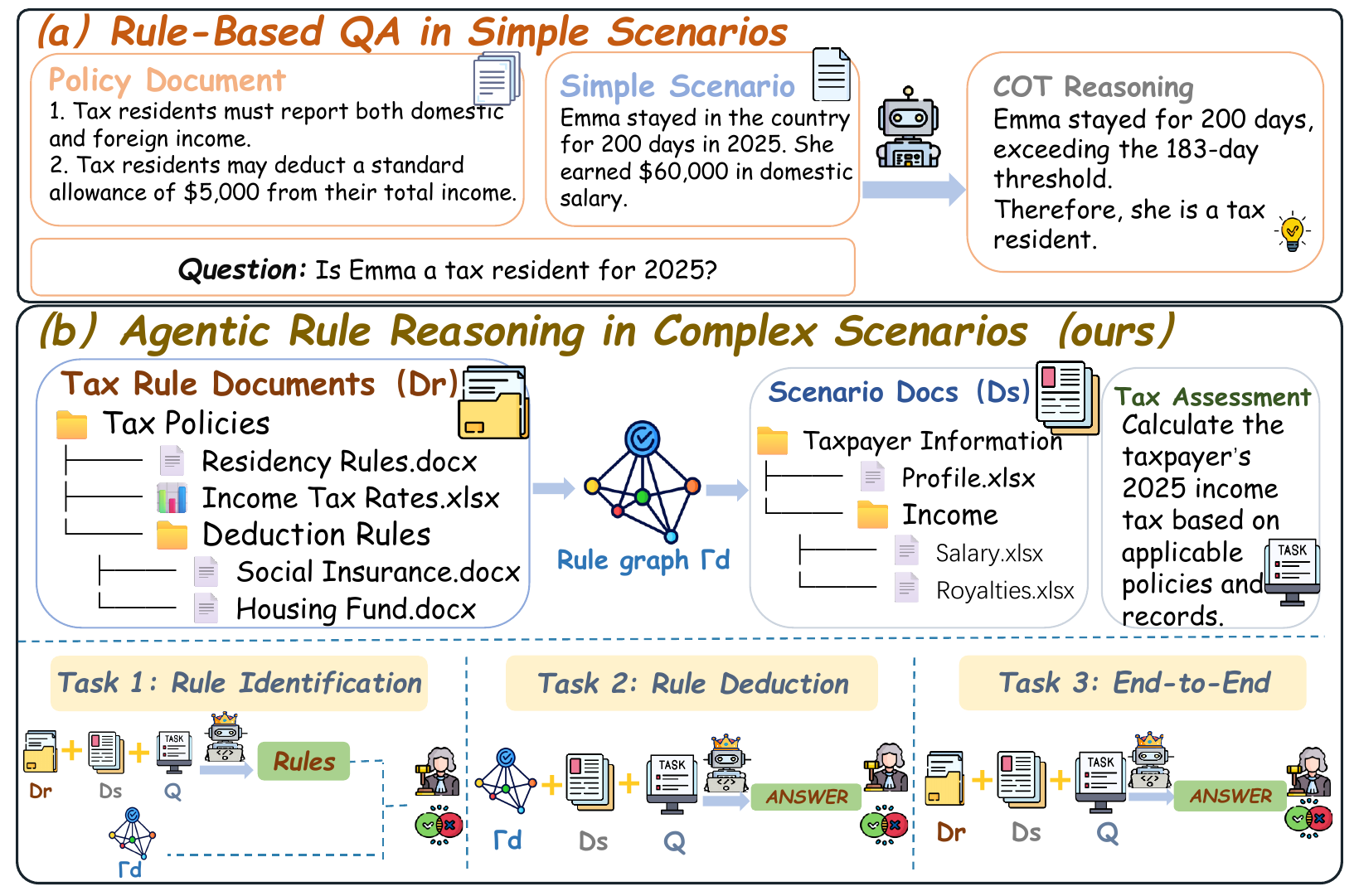}
    \caption{
Comparison of rule reasoning in prior work
and complex real-world scenarios.
Prior work typically presents scenarios
using a few factual statements, whereas
RuleTrek requires agents to integrate complex
scenario information distributed across
heterogeneous files, identify applicable
rules from complex rule documents,
and apply them to derive correct decisions.
}
    \label{fig:motivating-example}
\end{figure}

Recent advances in large language models (LLMs)
have enhanced agents' capabilities in reasoning,
information retrieval, and tool use
\citep{openai2026gpt56,anthropic2026opus55,
google2026gemini31pro,glm5team2026glm5}.
These advances have enabled agents to tackle
increasingly complex tasks in specialized domains,
where solutions often depend on domain-specific
rules, such as tax residency rules for tax assessment.  As scenarios grow more complex, agents face increasing challenges in identifying applicable
rules from relevant documents and correctly
applying them to solve problems using
distributed scenario information. For example, tax assessment requires agents to identify and apply rules from regulatory
documents. As illustrated in Figure~\ref{fig:motivating-example}(b),
this becomes challenging when complex
scenario information such as taxpayer profiles,
salary records and royalty income is
distributed across multiple files.
Agents must integrate this information to identify
applicable rules concerning residency requirements,
tax rates, and deductions, and correctly apply
them to calculate tax liability.

However, existing benchmarks \citep{clark2020transformerssoftreasonerslanguage,sun2024instructionfollowingevaluatinginferential,zhu-etal-2025-tableeval,zhou-etal-2025-rulearena,yu2026factualknowledgebenchmarkinglearning,tang2026workspacebench10benchmarkingai,liang2026stageclawautomatedstatebasedagent} provide limited
evaluation of agents' ability to identify
and apply rules to solve problems in
complex and multi-file scenarios.
As illustrated in Figure~\ref{fig:motivating-example}(a),
rule-reasoning benchmarks
\citep{clark2020transformerssoftreasonerslanguage,
tafjord-etal-2021-proofwriter,
zhou-etal-2025-rulearena,
yu2026ruleweaverbenchmarkingrulecenteredscenario}
typically represent scenarios using only
a few factual statements, simplifying
the identification and application
of relevant rules.
Agent benchmarks in realistic
workspaces
\citep{xu2025theagentcompanybenchmarkingllmagents,
malay2026enterpriseopsgymenvironmentsevaluationsstateful,
guo2026agoraarchivegroundedbenchmarkagentic,
tang2026workspacebench10benchmarkingai}
evaluate agents' abilities to retrieve
distributed information, use tools, and
complete multi-step tasks, but rarely
evaluate their ability to identify and apply
domain-specific rules to solve problems
in complex scenarios.

To address this gap, we introduce
\textbf{RuleTrek}, a multi-domain benchmark
for evaluating agents' ability to reason
with rules in complex scenarios. In RuleTrek, agents operate in interactive
workspaces, where they must iteratively
explore rule documents and scenario files,
retrieve relevant information, and use
tools as needed to solve problems. 
RuleTrek covers six domains: taxation,
PostgreSQL, cybersecurity, software policy,
procurement, and game rules.
To construct complex scenarios that can be
solved using rules from a given rule repository,
we develop a \textbf{rule-grounded scenario
construction pipeline} consisting of three stages:
(1) extracting structured rule graphs from
real-world rule repositories to capture
domain-specific rules and their dependencies;
(2) generating a sub-scenario for each rule
in the graph, with each sub-scenario requiring
the application of its corresponding rule;
and (3) incrementally composing these
sub-scenarios according to rule dependencies,
while verifying consistency and solvability
at each composition step.

To systematically evaluate agentic rule reasoning
in complex scenarios, RuleTrek assesses three
capabilities:
(1) \textbf{Rule Identification (RI)} evaluates
agents' ability to identify suitable rules
and their dependencies from original rule
documents;
(2) \textbf{Rule Deduction (RD)} evaluates
agents' ability to integrate distributed
scenario information and correctly apply
the provided necessary rules to reason
over complex scenarios;
and (3) \textbf{End-to-End Rule Reasoning (E2E)}
jointly evaluates rule identification
and rule deduction, requiring agents to
integrate distributed information across
complex scenarios to identify and apply
applicable rules from original rule documents
to solve complex tasks.
Experiments on 11 frontier LLMs show that
the highest RI and RD scores reach only
62.47\% and 60.43\%, respectively,
while E2E scores range from 26.97\% to 55.29\%.
These results highlight the substantial
challenges of both identifying applicable
rules and correctly applying them to
complex scenarios with distributed information.

The main contributions of this paper
are summarized as follows:

\begin{itemize}

    \item To systematically evaluate agents'
capabilities in rule identification,
rule deduction, and end-to-end rule reasoning
in complex scenarios, this paper introduces
\textbf{RuleTrek}, a multi-domain benchmark
featuring complex, multi-file scenarios
that require agents to integrate distributed
information and correctly identify and
apply domain-specific rules.

    \item To construct complex scenarios that
can be solved using rules from given
rule documents, this paper proposes a
rule-grounded scenario construction
pipeline. The pipeline extracts rule graphs,
generates rule-specific sub-scenarios,
and incrementally composes them while
ensuring scenario consistency and
rule-based solvability.

    \item This paper evaluates 11 frontier LLMs
    on RuleTrek across three evaluation settings.
    The results reveal substantial challenges
    in identifying applicable rules and
    correctly applying them to complex
    scenarios with distributed information.

\end{itemize}

\begin{table*}[t]
\centering

\begin{minipage}[c]{0.67\linewidth}
\vspace{0pt}
\centering
\scriptsize
\setlength{\tabcolsep}{2.2pt}
\renewcommand{\arraystretch}{1.10}

\resizebox{\linewidth}{!}{%
\begin{tabular}{lccccc}
\toprule

\textbf{Benchmark}
& \textbf{Rule App.}
& \textbf{Native Rules}
& \textbf{Dist. Scenario}
& \textbf{Multi-Domain}
& \textbf{Workspace} \\

\midrule

\multicolumn{6}{l}{\textbf{Rule-Reasoning Benchmarks}} \\
\midrule

RuleTaker
\citep{clark2020transformerssoftreasonerslanguage}
& \pmark & \xmark & \xmark & \xmark & \xmark \\

PolicyQA
\citep{ahmad-etal-2020-policyqa}
& \xmark & \pmark & \xmark & \xmark & \xmark \\

LegalBench
\citep{guha2023legalbenchcollaborativelybuiltbenchmark}
& \pmark & \pmark & \xmark & \pmark & \xmark \\

RuleArena
\citep{zhou-etal-2025-rulearena}
& \cmark & \pmark & \xmark & \cmark & \xmark \\

RuleWeaver
\citep{yu2026ruleweaverbenchmarkingrulecenteredscenario}
& \cmark & \xmark & \xmark & \cmark & \xmark \\

SearchFireSafety
\citep{chae2026caselawevaluatingstructureaware}
& \pmark & \pmark & \xmark & \xmark & \xmark \\

\midrule

\multicolumn{6}{l}{\textbf{Workspace Benchmarks}} \\
\midrule

TheAgentCompany
\citep{xu2025theagentcompanybenchmarkingllmagents}
& \xmark & \xmark & \pmark & \pmark & \cmark \\

EnterpriseOps-Gym
\citep{malay2026enterpriseopsgymenvironmentsevaluationsstateful}
& \pmark & \xmark & \pmark & \cmark & \cmark \\

AGORA
\citep{guo2026agoraarchivegroundedbenchmarkagentic}
& \xmark & \xmark & \cmark & \cmark & \cmark \\

Workspace-Bench
\citep{tang2026workspacebench10benchmarkingai}
& \xmark & \xmark & \cmark & \cmark & \cmark \\

PhantomPolicy
\citep{wu2026policyinvisibleviolationsllmbasedagents}
& \pmark & \xmark & \pmark & \xmark & \cmark \\

\midrule

\textbf{RuleTrek (Ours)}
& \cmark
& \cmark
& \cmark
& \cmark
& \cmark \\

\bottomrule
\end{tabular}%
}

\end{minipage}%
\hspace{0.01\linewidth}%
\begin{minipage}[c]{0.32\linewidth}
\vspace{0pt}
\centering

\includegraphics[
    width=\linewidth,
    height=\linewidth,
    keepaspectratio
]{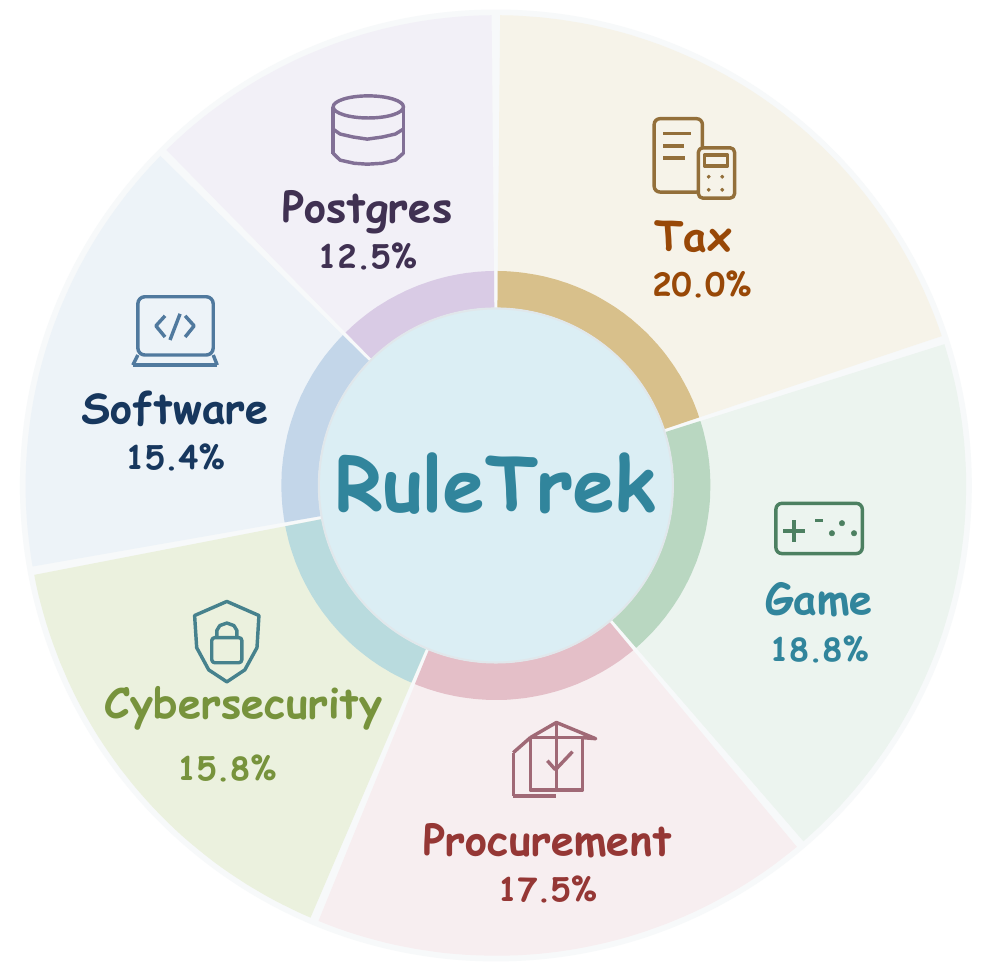}

\end{minipage}%

\vspace{5pt}

\caption{
\textbf{Left:} Comparison of RuleTrek with representative
rule-reasoning and workspace benchmarks.
\textbf{Rule App.} indicates whether tasks require
applying domain rules to concrete cases;
\textbf{Native Rules} whether rules retain their
original representations;
\textbf{Dist. Scenario} whether scenario evidence
is distributed across multiple artifacts;
\textbf{Multi-Domain} whether the benchmark covers
multiple application domains; and
\textbf{Workspace} whether agents operate in an
interactive environment.
\cmark, \xmark, and \pmark denote full, absent,
and partial support, respectively.
\textbf{Right:} Distribution of RuleTrek instances
across six domains.
}

\label{tab:benchmark-comparison}
\end{table*}

\section{Related Work}
\label{sec:related-work}

Agentic rule reasoning in complex scenarios
is closely related to two lines of research:
rule-reasoning benchmarks and agent benchmarks
in realistic workspaces.
Table~\ref{tab:benchmark-comparison}
compares representative benchmarks
from both lines.

\paragraph{Rule-Reasoning Benchmarks.}
Rule-reasoning benchmarks evaluate models'
ability to identify and apply rules
to scenario information.
RuleTaker studies deduction over normalized
rules and facts
\citep{clark2020transformerssoftreasonerslanguage},
while PolicyQA and LegalBench extend rule reasoning
to real-world policies and legal domains
\citep{ahmad-etal-2020-policyqa,
guha2023legalbenchcollaborativelybuiltbenchmark}.
More recent benchmarks introduce more complex
and realistic rule systems.
RuleArena evaluates reasoning over complex
real-world rules
\citep{zhou-etal-2025-rulearena},
RuleWeaver constructs scenario-based reasoning
tasks from corpus-derived rules
\citep{yu2026ruleweaverbenchmarkingrulecenteredscenario},
and SearchFireSafety evaluates retrieval
and reasoning over structured statutory sources
\citep{chae2026caselawevaluatingstructureaware}.
However, even when these benchmarks provide
original rule repositories, they typically
use simplified scenarios rather than scenarios
with distributed information, substantially
reducing the difficulty of identifying
and applying applicable rules.

\paragraph{Agent Benchmarks in Realistic Workspaces.}
Agent benchmarks in realistic workspaces
evaluate agents' abilities to retrieve
distributed information, use tools,
and complete multi-step tasks.
TheAgentCompany evaluates agents on
realistic workplace tasks
\citep{xu2025theagentcompanybenchmarkingllmagents},
while EnterpriseOps-Gym focuses on
stateful enterprise operations
\citep{malay2026enterpriseopsgymenvironmentsevaluationsstateful}.
AGORA evaluates reasoning over workplace
document archives
\citep{guo2026agoraarchivegroundedbenchmarkagentic},
and Workspace-Bench examines tasks involving
large-scale file dependencies
\citep{tang2026workspacebench10benchmarkingai}.
PhantomPolicy further investigates
policy-sensitive agent behavior
\citep{wu2026policyinvisibleviolationsllmbasedagents}.
Although these benchmarks evaluate agents'
ability to complete tasks in realistic
workspaces or comply with behavioral policies,
they rarely assess agents' ability to
identify and apply domain-specific rules
to solve problems in complex scenarios.

RuleTrek addresses this gap by evaluating
rule identification, rule deduction,
and end-to-end rule reasoning in complex,
multi-file scenarios.
It requires agents to integrate distributed
scenario information and reason over
applicable rules in interactive workspaces.
\section{RuleTrek Construction}
\label{sec:construction}

\begin{figure*}[t]
    \centering
    \includegraphics[width=\textwidth]{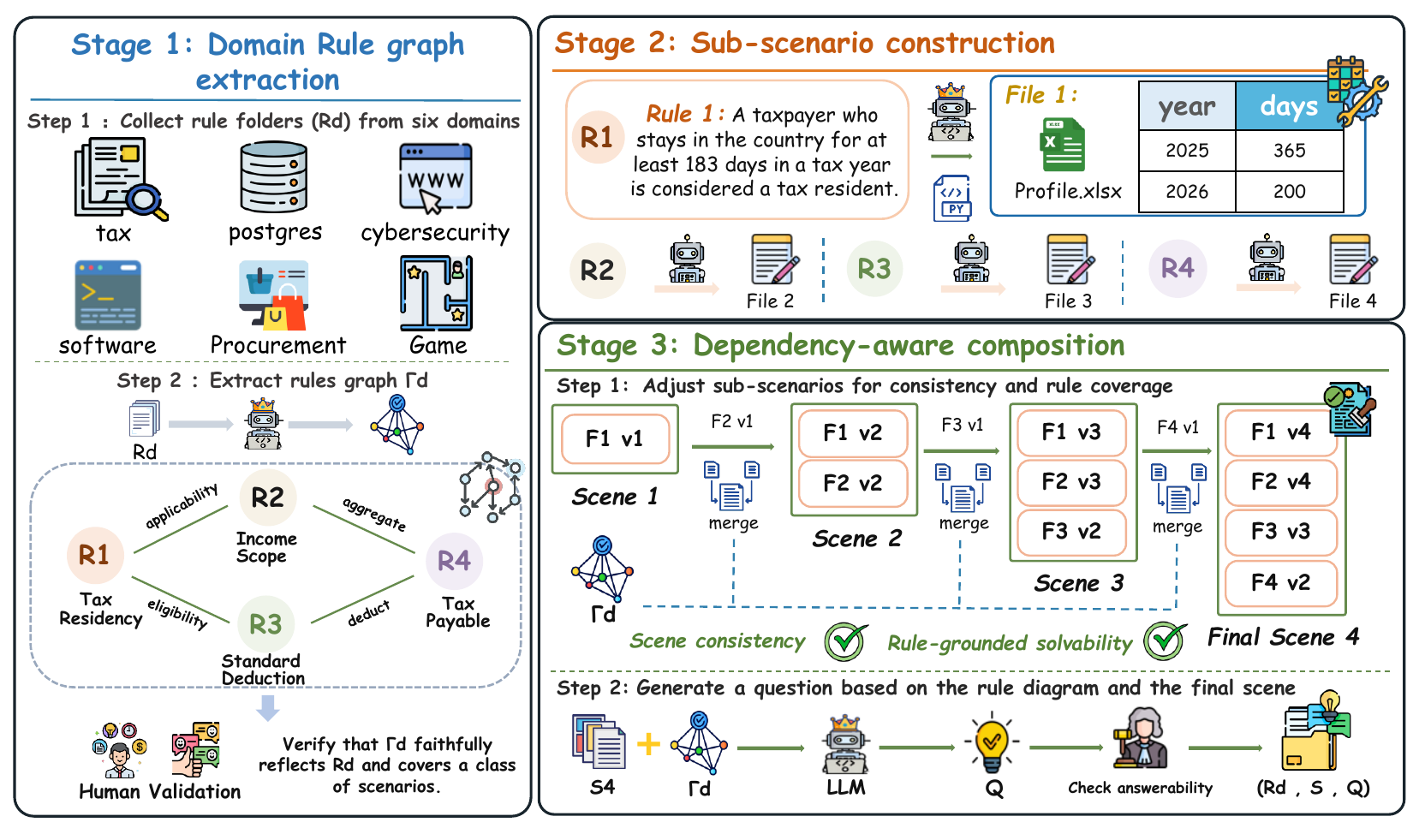}
    \caption{
    Overview of the RuleTrek construction pipeline.
    Stage 1 extracts and verifies rule graphs
    from real-world rule repositories.
    Stage 2 generates rule-specific sub-scenarios
    requiring the application of corresponding rules.
    Stage 3 incrementally composes sub-scenarios
    according to rule dependencies, verifies
    consistency and solvability at each step,
    and generates questions from the final scenarios.
    }
    \label{fig:overview}
\end{figure*}

RuleTrek constructs complex scenarios
grounded in real-world rule repositories.
As illustrated in Figure~\ref{fig:overview},
we develop a \textbf{rule-grounded scenario
construction pipeline} consisting of three stages:
(1) extracting and verifying structured rule graphs;
(2) generating rule-specific sub-scenarios
that require applying their corresponding rules;
(3) incrementally composing these sub-scenarios
according to rule dependencies, with consistency
and solvability checks at each step. All LLM-based construction steps use
GPT-6-Astra~\cite{openai2026modeldocs}.

\subsection{Stage 1: Domain Rule Graph Extraction}
\label{sec:rule-graph-extraction}

For each instance in domain $d$, RuleTrek
collects a set of relevant real-world rule files
$\mathcal{R}_d$.
The repositories cover six diverse domains:
taxation, PostgreSQL, cybersecurity,
software policy, procurement, and game rules.
They preserve their original organization
and formats, including documents, tables,
configuration files, and source code.
RuleTrek employs an LLM to extract a
structured rule graph
$\Gamma_d=(V_d,E_d)$ from $\mathcal{R}_d$.
Each node $R_i\in V_d$ represents a rule
grounded in the original repository,
while edges in $E_d$ capture explicit relationships
between rules, including applicability
dependencies, eligibility conditions,
aggregation, exceptions, overrides,
precedence, and temporal constraints.
The extracted graph is manually verified
against the original repository.
Human experts carefully examine the correctness
of extracted rules, source references,
and rule dependencies, and revise any
inaccurate or missing information.
The verified graph $\Gamma_d$ serves
as the reliable rule structure for subsequent
scenario construction of the corresponding instance.

\subsection{Stage 2: Rule-Specific Sub-Scenario Construction}
\label{sec:micro-scenario}

Directly generating a complex scenario
from the rule graph makes it difficult
to control which rules are exercised
and whether the scenario remains consistent.
RuleTrek constructs sub-scenarios
for individual rules.
For each rule $R_i\in V_d$, an LLM generates
a corresponding sub-scenario $s_i$
based on $R_i$ and its dependencies
in $\Gamma_d$.
Each sub-scenario provides the information
needed to apply $R_i$, and resolving it
requires that rule.
This approach controls which rules are
exercised during construction.
Sub-scenarios represent information
through files commonly used in the
corresponding domain.
For example, a tax-residency rule may
be instantiated through a spreadsheet
recording a taxpayer's days of residence
across different tax years.
Other domains may involve configuration
files, logs, forms, messages, tables,
or source code.
These files provide entities, events,
and temporal or state information
needed for rule application.
Each sub-scenario is manually reviewed for
rule consistency, solvability, and sufficient
information for rule application.
Experts remove redundant cases and ensure
diversity in entities, contexts,
and information organization.

\subsection{Stage 3: Dependency-Aware Composition}
\label{sec:scenario-composition}

RuleTrek employs an LLM to incrementally
compose the generated sub-scenarios
according to the dependencies encoded
in $\Gamma_d$.
Starting from an initial scenario $S_1$,
additional sub-scenarios are systematically integrated
to produce the final complex scenario $S_T$.
At each step, the composition operation is
defined as
\begin{equation}
    S_{t+1} =
    \mathcal{C}_{\Gamma_d}(S_t,s_{i_{t+1}}),
\end{equation}
where $s_{i_{t+1}}$ is the newly introduced
sub-scenario and $\mathcal{C}_{\Gamma_d}$
denotes LLM-assisted composition guided
by rule dependencies.
Composition involves more than combining files.
Introducing a new sub-scenario may require
revising existing files to align entities,
timestamps, states, versions, units,
and cross-file references.
These necessary revisions ensure that information
from different sub-scenarios forms a
coherent scenario in which multiple
rules interact.
After each composition step, human experts
carefully verify the consistency and solvability
of the intermediate scenario under
the verified rule graph $\Gamma_d$.
Consistency checks examine whether information
remains coherent across files, while
solvability checks determine whether
the scenario can be resolved using
the underlying rules.
If either check fails, the LLM revises
the affected files, followed by further
human verification.
Composition proceeds only after
both checks are satisfied.
The resulting scenario $S_T$ contains
interacting rules and information
distributed across multiple files.
Throughout this iterative process, $\Gamma_d$
remains unchanged, allowing RuleTrek
to increase scenario complexity
without modifying the original underlying rules.

\subsection{Question Instantiation}
\label{sec:question-instantiation}

After the final scenario $S_T$ passes
consistency and solvability checks,
RuleTrek generates a question $Q$
and its answer $A$ based on $S_T$ and $\Gamma_d$.
Human experts solve each question using
the original rules and scenario files,
verifying answerability, answer uniqueness,
and rule compliance.
Invalid instances are revised and revalidated.
Each benchmark instance comprises
the rule repository $\mathcal{R}_d$,
verified graph $\Gamma_d$, scenario $S_T$,
and question $Q$, which jointly support
evaluation of the three core capabilities.

\subsection{Three Evaluation Capabilities}
\label{sec:tasks}

RuleTrek evaluates three capabilities
using the same complex scenarios and
questions in interactive workspaces.
Agents can search and inspect files
and execute code as needed.
The evaluation settings differ in
the rule information provided
and the required outputs.

\paragraph{Rule Identification (RI).}
RI evaluates agents' ability to identify
applicable rules and their dependencies
from original rule documents.
Given $(\mathcal{R}_d,S_T,Q)$,
agents must identify the rules needed
to solve the problem and their dependencies,
producing an estimated rule graph
$\widehat{\Gamma}$.
The verified graph $\Gamma_d$ is not provided,
and the final answer to $Q$ is not required.

\paragraph{Rule Deduction (RD).}
RD evaluates agents' ability to correctly
apply provided rules to complex scenarios.
Given $(\Gamma_d,S_T,Q)$,
agents must integrate distributed scenario
information, evaluate rule conditions,
resolve dependencies and exceptions,
and derive the final answer $\widehat{A}$.
The complete verified rule graph $\Gamma_d$
is provided, allowing RD to evaluate
rule application without requiring
rule identification from original documents.

\paragraph{End-to-End Rule Reasoning (E2E).}
E2E evaluates agents' ability to jointly
identify and apply rules to solve
complex tasks.
Given $(\mathcal{R}_d,S_T,Q)$,
agents must integrate distributed scenario
information, identify applicable rules
and their dependencies from original
rule documents, and correctly apply
them to produce the final answer
$\widehat{A}$.
\section{Experiments}
\label{sec:experiments}
In this section, we evaluate 11 frontier LLMs on RuleTrek.
We first describe the experimental setup and report the
main results across three evaluation tasks and six domains.
We then analyze the impact of rule complexity, agent
interaction efficiency, and the challenges of rule
identification and application.
\subsection{Experimental Setup}
\label{sec:system-setup}

\paragraph{Evaluated models.}
We evaluate 11 frontier LLMs from seven model families:
GPT-6.1-Sol, GPT-5.5, and GPT-6-Luna~\cite{openai2026modeldocs};
Grok-4.7~\cite{xai2026modeldocs};
DeepSeek-V4-Pro and DeepSeek-V4.1-Flash~\cite{deepseek2026modeldocs};
Qwen3.8-Flash~\cite{alibaba2026modeldocs};
GLM-5.3-Flash~\cite{zai2026glm53flash};
Gemini-3.7-Flash and Gemini-2.5-Flash~\cite{google2026geminimodeldocs};
and HY3~\cite{tencent2026hunyuanmodeldocs}.

\begin{table*}[t]

\definecolor{maintablebest}{RGB}{228,246,228}
\definecolor{maintablesecond}{RGB}{252,232,232}

\vspace{-6pt}

\centering
\scriptsize

\setlength{\tabcolsep}{2.0pt}
\renewcommand{\arraystretch}{1.25}
\setlength{\aboverulesep}{0pt}
\setlength{\belowrulesep}{0pt}

\newcommand{\bestcell}[1]{%
  \cellcolor{maintablebest}\textbf{#1}}
\newcommand{\secondcell}[1]{%
  \cellcolor{maintablesecond}#1}

\resizebox{\textwidth}{!}{%
\begin{tabular}{l|ccc|ccc|ccc|ccc|ccc|ccc|ccc}

\toprule

\multirow[c]{2}{*}{\textbf{Model}}
& \multicolumn{3}{c|}{\textbf{Avg.}}
& \multicolumn{3}{c|}{\textbf{Tax}}
& \multicolumn{3}{c|}{\textbf{Game}}
& \multicolumn{3}{c|}{\textbf{Procurement}}
& \multicolumn{3}{c|}{\textbf{Cybersecurity}}
& \multicolumn{3}{c|}{\textbf{Software Policy}}
& \multicolumn{3}{c}{\textbf{PostgreSQL}}
\\

&
\textbf{RI} & \textbf{RD} & \textbf{E2E}
& \textbf{RI} & \textbf{RD} & \textbf{E2E}
& \textbf{RI} & \textbf{RD} & \textbf{E2E}
& \textbf{RI} & \textbf{RD} & \textbf{E2E}
& \textbf{RI} & \textbf{RD} & \textbf{E2E}
& \textbf{RI} & \textbf{RD} & \textbf{E2E}
& \textbf{RI} & \textbf{RD} & \textbf{E2E}
\\

\midrule

GPT-6.1-Sol
& \bestcell{62.47} & \bestcell{60.43} & \bestcell{55.29}
& \secondcell{65.44} & 86.42 & \secondcell{85.89}
& 67.76 & \bestcell{69.20} & \bestcell{71.99}
& \bestcell{57.78} & \bestcell{43.14} & \bestcell{26.92}
& 64.81 & \bestcell{56.78} & \secondcell{47.82}
& \bestcell{67.39} & \bestcell{68.00} & \bestcell{59.31}
& 47.28 & 25.17 & \secondcell{25.49}
\\

Grok-4.7
& \secondcell{61.86} & 56.12 & \secondcell{51.34}
& \bestcell{65.76} & 87.14 & \bestcell{85.95}
& 67.59 & \secondcell{68.75} & 65.53
& \secondcell{56.48} & \secondcell{30.86} & 20.40
& 68.57 & \secondcell{48.11} & \bestcell{48.15}
& \secondcell{60.03} & 61.82 & \secondcell{57.01}
& \secondcell{48.33} & 25.99 & 15.02
\\

DeepSeek-V4.1-Flash
& 59.21 & \secondcell{56.39} & 50.06
& 64.27 & \bestcell{92.30} & 77.78
& 64.60 & 66.57 & 62.93
& 45.83 & 28.00 & 23.62
& \bestcell{72.60} & 42.80 & 41.98
& 56.69 & \secondcell{65.52} & 55.91
& 47.93 & \secondcell{29.41} & \bestcell{26.46}
\\

DeepSeek-V4-Pro
& 58.25 & 51.81 & 49.07
& 64.78 & \secondcell{88.13} & 82.40
& \secondcell{68.42} & 65.91 & \secondcell{67.93}
& 48.30 & 23.47 & \secondcell{25.16}
& 66.50 & 43.18 & 40.83
& 50.59 & 49.82 & 46.62
& 45.43 & 25.59 & 14.37
\\

GLM-5.3-Flash
& 56.53 & 51.06 & 47.16
& 58.94 & 79.90 & 81.00
& 60.23 & 66.78 & 63.75
& 47.06 & 21.29 & 19.63
& \secondcell{72.00} & 42.97 & 36.85
& 51.13 & 59.98 & 53.14
& 47.43 & 22.25 & 12.34
\\

Gemini-3.7-Flash
& 55.27 & 50.05 & 46.26
& 55.30 & 81.67 & 71.87
& 59.70 & 57.28 & 54.46
& 45.02 & 27.83 & 24.70
& 67.89 & 42.55 & 43.20
& 54.13 & 58.25 & 48.90
& \bestcell{48.34} & 19.12 & 23.76
\\

GPT-5.5
& 60.20 & 52.57 & 45.47
& 62.49 & 75.73 & 61.58
& \bestcell{71.83} & 65.86 & 61.70
& 54.40 & 28.09 & 21.96
& 68.66 & 42.41 & 47.42
& 56.09 & 61.54 & 52.50
& 41.53 & \bestcell{31.65} & 17.14
\\

Qwen3.8-Flash
& 55.02 & 52.01 & 44.59
& 60.10 & 83.81 & 70.55
& 60.51 & 64.01 & 66.14
& 46.65 & 26.05 & 19.98
& 59.62 & 43.50 & 32.98
& 53.48 & 60.56 & 48.42
& 46.39 & 19.74 & 15.15
\\

HY3
& 50.97 & 47.62 & 39.47
& 54.80 & 85.84 & 62.45
& 59.09 & 62.74 & 60.46
& 42.27 & 18.46 & 15.45
& 60.17 & 34.15 & 32.98
& 44.42 & 52.20 & 42.79
& 41.23 & 16.03 & 8.96
\\

GPT-6-Luna
& 50.86 & 44.46 & 34.36
& 48.41 & 59.81 & 43.31
& 59.52 & 58.58 & 55.72
& 47.09 & 27.33 & 18.00
& 53.81 & 36.39 & 30.09
& 51.06 & 55.32 & 41.01
& 43.08 & 19.57 & 8.08
\\

Gemini-2.5-Flash
& 33.35 & 35.26 & 26.97
& 36.24 & 36.52 & 39.19
& 52.50 & 60.19 & 34.60
& 40.52 & 27.39 & 14.45
& 27.37 & 24.73 & 14.70
& 23.92 & 42.28 & 41.03
& 9.21 & 11.55 & 11.66
\\

\bottomrule
\end{tabular}%
}
\caption{Main results (\%) across six domains and three tasks:
Rule Identification (RI), Rule Deduction (RD), and
End-to-End Decision (E2E).
\protect\colorbox{maintablebest}{Green} and
\protect\colorbox{maintablesecond}{red} cells indicate the best and
second-best scores, respectively; best scores are bold.}
\label{tab:main_results_240}
\vspace{-10pt}
\end{table*}

\paragraph{Evaluation set.}
RuleTrek contains 240 instances across six domains, with
30--48 instances per domain
(Table~\ref{tab:dataset-statistics}).
The rule repositories are drawn from cybersecurity
standards and policies~\cite{nist2026oscalcontent,falco2026prempti},
game rules~\cite{downfallx2026srd,openttd2026wikidata},
PostgreSQL documentation and tools~\cite{postgresql2026docs,pgbackrest2026repository},
procurement standards~\cite{ocds2026standard,ocds2026extensions},
software policies~\cite{kyverno2026policies,opa2026gatekeeper,github2026sitepolicy},
and tax computation systems~\cite{irs2026withholding,psl2026taxcalculator,psl2026costcapital}. All source repositories were selected under licenses or terms permitting reproduction, modification, and redistribution for academic research.
The operative rule graphs contain an average of
9.05 nodes and 8.69 edges, with a mean depth of 2.69.
Graph depth is measured as the longest directed path
in edges.

\begin{table}[t]
\centering
\footnotesize
\setlength{\tabcolsep}{3pt}
\renewcommand{\arraystretch}{1.08}
\begin{tabular}{lrrrr}
\toprule
Domain & $N$ & Nodes & Edges & Depth \\
\midrule
Cybersecurity & 38 & 7.05 & 6.18 & 1.55 \\
Game & 45 & 5.56 & 4.91 & 1.82 \\
PostgreSQL & 30 & 13.90 & 13.07 & 4.10 \\
Procurement & 42 & 8.74 & 7.55 & 2.43 \\
Software & 37 & 7.68 & 7.84 & 2.78 \\
Tax & 48 & 12.21 & 13.15 & 3.67 \\
\midrule
\textbf{All} & \textbf{240} & \textbf{9.05} & \textbf{8.69} & \textbf{2.69} \\
\bottomrule
\end{tabular}
\caption{Evaluation-set statistics. $N$: number of instances; Nodes/Edges: mean operative rule-graph size; Depth: mean longest directed path length in edges. All means are computed over instances.}
\label{tab:dataset-statistics}
\end{table}

\definecolor{bestgreen}{RGB}{220,238,220}
\definecolor{secondred}{RGB}{248,226,226}

\newcommand{\best}[1]{\cellcolor{bestgreen}\textbf{#1}}
\newcommand{\second}[1]{\cellcolor{secondred}#1}

\paragraph{Evaluation details.}
All models are evaluated using pi CLI~\cite{zechner2026pi}
in isolated workspaces with file and shell access,
without external network access.
Each run has a 30-minute time limit and a maximum
of 200 agent steps, and produces a JSON submission.
Each model is run three times at temperature 1,
and scores are averaged across runs.
We use gpt-5.6-luna ~\cite{openai2026gpt56luna} at temperature 0
as the evaluation judge.
For RI, the judge evaluates the correctness of
the identified rule set and the logical validity
of the stated relationships between rules.
For RD and E2E, the judge assesses answer correctness
and assigns a reasoning score based on four dimensions:
fact grounding, rule application, computation,
and evidential support.
The final score is the product of binary answer
correctness (0 or 1) and the reasoning score.
All scores are reported on a 0--100 scale and
averaged over 240 instances, with failed or
missing submissions assigned zero.

\subsection{Main Results}

Table~\ref{tab:main_results_240} reports
the performance of 11 frontier LLMs
on RI, RD, and E2E across six domains.

\paragraph{Identifying applicable rules remains challenging.}
Agents struggle to accurately identify applicable
rules and their dependencies from
original rule repositories.
GPT-6.1-Sol achieves the highest
average RI score of 62.47\%,
followed by Grok-4.7 at 61.86\%
and GPT-5.5 at 60.20\%.
Other models achieve relatively lower scores,
with Gemini-2.5-Flash obtaining only 33.35\%.
These results indicate substantial
and persistent limitations in identifying applicable
rules and their dependencies from
original rule documents.

\paragraph{Applying rules in complex scenarios remains challenging.}
Even when the verified rule graph is provided,
the highest average RD score reaches only
60.43\%, achieved by GPT-6.1-Sol,
followed by DeepSeek-V4.1-Flash at 56.39\%
and Grok-4.7 at 56.12\%.
When agents must also identify applicable
rules from original documents,
E2E scores range from 26.97\% to 55.29\%.
Overall, these scores show that agents
struggle to integrate distributed scenario
information and correctly apply rules,
even when the verified rules are provided.

\begin{table}[h]
\centering
\footnotesize
\setlength{\tabcolsep}{5pt}
\renewcommand{\arraystretch}{1.15}
\begin{tabularx}{\linewidth}{@{}lX@{}}
\toprule
\textbf{Instance} & \textbf{Missed or Misidentified Rules} \\
\midrule

Cyber-007 &
Fine-grained firewall and remote-access controls. \\

\addlinespace[3pt]

Tax-007 &
Data-loading and tax-year initialization rules. \\

\addlinespace[3pt]

Game-042 &
Distinct rules for different travel paces. \\

\bottomrule
\end{tabularx}
\caption{
Examples of frequently missed or misidentified
rules in Rule Identification (RI) across models.
}
\label{tab:heterogeneity_cases}
\end{table}

\subsection{Error Analysis of Rule Identification}

We analyze common rule identification errors across models, as shown in Table~\ref{tab:heterogeneity_cases}. In Cybersecurity, models often overlook condition-specific controls despite identifying general requirements. In Tax, they miss necessary supporting rules for data loading and tax-year initialization. In Game, they incorrectly merge movement rules for different travel paces with distinct applicability conditions. These errors highlight three key challenges: mapping scenario evidence to specific rules, identifying rule dependencies, and distinguishing rules with different applicability conditions. Overall, complex scenarios make it difficult for agents to identify complete and correctly structured sets of applicable rules.

\begin{figure}[h]
    \centering
    \includegraphics[width=\linewidth]{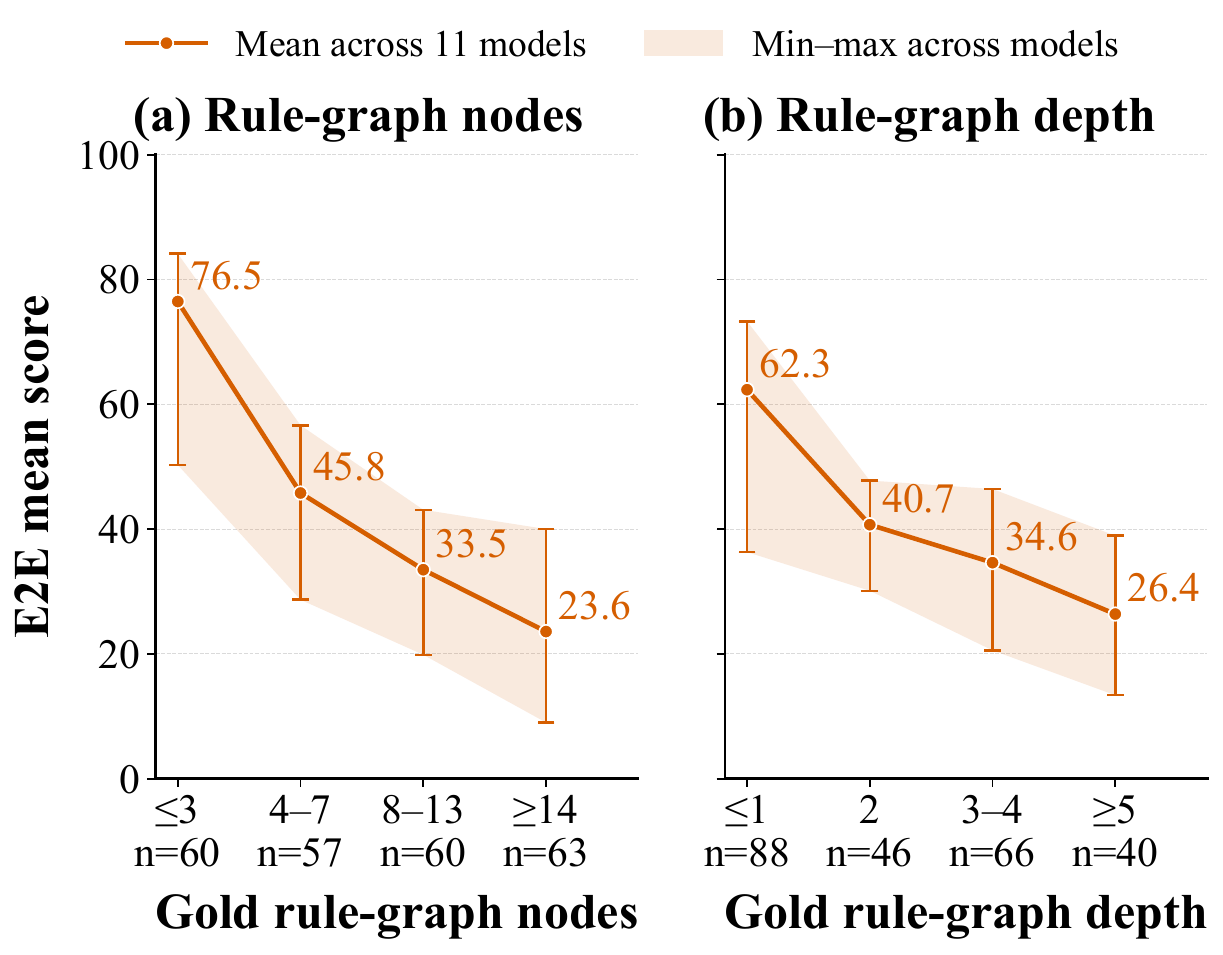}
  \caption{
Rule-graph complexity and E2E performance.
Tasks are grouped by (a) rule-node count and
(b) rule-graph depth.
Lines show mean scores across 11 models;
shaded bands and error bars indicate the
minimum--maximum range.
$n$ denotes the number of tasks per bin.
}
    \label{fig:depth}
\end{figure}
\subsection{Performance Degrades with Rule-Graph Complexity}
\label{sec:depth-analysis}

We further examine the relationship between
rule-graph complexity and model performance
from two perspectives: the number of rule
nodes and rule-graph depth.
As shown in Figure~\ref{fig:depth},
E2E performance consistently decreases
as either measure increases.
The mean score across 11 models drops
from 76.5 to 23.6 as the number of rule
nodes increases from at most 3 to at least 14.
Similarly, the score decreases from 62.3
to 26.4 as rule-graph depth increases
from at most 1 to at least 5.
These trends suggest that agents face
greater difficulties when reasoning over
larger sets of interacting rules and
longer dependency chains.
More rule nodes increase the number of
rules that agents must identify and
consider, while greater graph depth
requires reasoning through more
interdependent rule applications.
Overall, increasing rule-graph complexity
is associated with greater challenges
in end-to-end rule reasoning over
complex scenarios.

\subsection{Complex Scenarios Challenge Both Rule Identification and Application}
\label{sec:rule-identification-bottleneck}

To quantify the difficulty of E2E rule reasoning, Figure 4 compares the average reasoning scores for RD and E2E across seven models.
In RD, agents are provided with the verified
rule graph, whereas E2E requires them to
identify applicable rules and their dependencies
from original rule documents.
RD consistently outperforms E2E by
3.79--8.15 points, showing that rule
identification introduces additional
difficulty in complex scenarios.
However, RD performance remains limited
despite access to the verified rule graph,
indicating that even with the correct rules,
agents struggle to integrate distributed
scenario information and correctly apply
interacting rules to solve complex problems.
\begin{figure}[h]
    \centering
    \includegraphics[width=\linewidth]{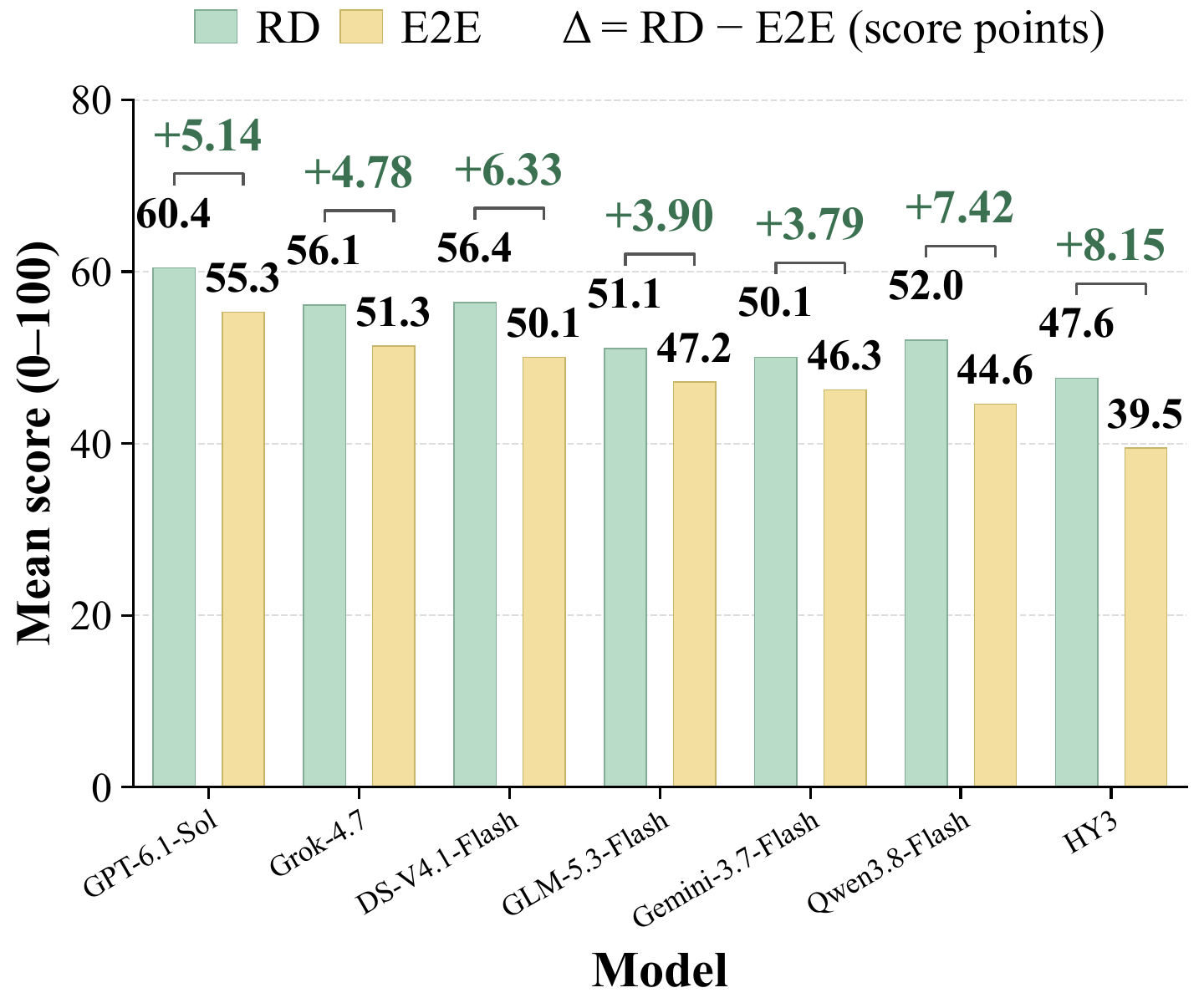}
    \caption{
Comparison of mean RD and E2E scores across seven models.
Annotations show $\Delta = \mathrm{RD} - \mathrm{E2E}$.
}
    \label{fig:gap}
\end{figure}
To further examine E2E failures, we classify
GPT-6.1-Sol's failed trajectories by their
earliest decisive error into four categories:
retrieval misses (missing relevant evidence),
rule-selection errors (incorrect rule sets
despite available evidence),
reasoning errors (incorrect rule application),
and final-synthesis errors (incorrect outputs).
Rule-selection errors account for 54.9\%
of failures, the largest category, indicating
that agents frequently struggle to identify
applicable rules even with relevant evidence.
Together with the limited RD performance,
these findings suggest that complex scenarios
challenge both rule identification and
application, jointly constraining
end-to-end reasoning performance.

\subsection{Model Differences in Agent Behavior and E2E Performance}
\label{sec:efficiency-analysis}

\begin{figure}[h]
    \centering
    \includegraphics[width=\linewidth]{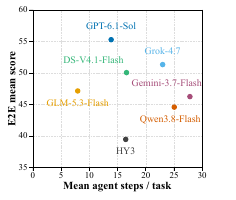}
    \caption{
    Relationship between mean agent steps per task
    and mean E2E scores across seven models.
    Each point represents one model.
    }
    \label{fig:efficiency}
\end{figure}

Figure~\ref{fig:efficiency} compares mean
E2E scores and agent steps across seven
representative models from different model families.
Models exhibit substantial differences
in both performance and interaction efficiency.
GPT-6.1-Sol achieves the highest E2E score
of 55.29\% with relatively few steps,
whereas Gemini-3.7-Flash and Qwen3.8-Flash
require considerably more steps but obtain
lower scores of 46.26\% and 44.59\%,
respectively.
However, fewer steps do not always
indicate better performance, as GLM-5.3-Flash
uses fewer steps than most models
but achieves an E2E score of only 47.16\%.
These observations suggest that successful
rule reasoning depends not simply on
the number of interactions, but on
how effectively agents identify relevant
information and apply rules throughout
the reasoning process.
Appendix~\ref{app:trajectory-analysis}
further illustrates these differences
through a game scenario involving
evolving states and interacting rules.
GPT-6.1-Sol correctly recognizes that
a protective ward is consumed upon
its first trigger and tracks the resulting
state changes, completing the task
in 34 steps.
In contrast, Qwen3.8-Flash retrieves
the same protection rule but incorrectly
treats the ward as still active.
This error propagates through subsequent
events, leading to an incorrect answer
after 57 steps.
The comparison demonstrates that
retrieving applicable rules is insufficient
without correctly applying them and
maintaining consistent scenario states.

\subsection{Scoring Reliability}
\label{sec:scoring_reliability}

\paragraph{Human Annotation and Agreement.}
We evaluate automatic scoring reliability through
independent human annotation of all instances
across RI, RD, and E2E.
Two annotators independently score each instance
using the same inputs, reference answers, and
criteria as the primary judge,
gpt-5.6-luna~\citep{openai2026gpt56luna}.
As shown in Table~\ref{tab:scoring_reliability},
human--judge Spearman correlations range from
0.822 to 0.857, compared with 0.933--0.951
for human--human agreement.
These results indicate substantial scoring agreement,
although human agreement remains higher.

\paragraph{Multi-family Judge Robustness.}
To avoid potential model-specific scoring biases, we further rescore the same submissions using
GLM-5~\citep{glm5team2026glm5} and
Gemini-2.5-Pro~\citep{google2025gemini25pro},
with identical evaluation inputs and criteria
and the primary judge as reference.
Submission-level Spearman correlations reach
0.903--0.934, and model-ranking correlations
reach 0.955--0.991.
These results suggest consistent scoring behavior
and relatively stable model rankings
across different judge families.

\begin{table}[t]
\centering
\footnotesize
\setlength{\tabcolsep}{3pt}
\renewcommand{\arraystretch}{1.08}

\begin{tabular*}{\columnwidth}
{@{\extracolsep{\fill}}llcc@{}}
\toprule
\multicolumn{4}{@{}l}{
\textit{(a) Human agreement}} \\
\midrule
\textbf{Task} & \textbf{Pair}
& \textbf{Spearman $\rho$}
& \textbf{Pearson $r$} \\
\midrule
RI  & H--J & 0.822 & 0.821 \\
    & H--H & 0.951 & 0.939 \\
\addlinespace[2pt]
RD  & H--J & 0.857 & 0.798 \\
    & H--H & 0.933 & 0.919 \\
\addlinespace[2pt]
E2E & H--J & 0.846 & 0.837 \\
    & H--H & 0.945 & 0.928 \\
\bottomrule
\end{tabular*}

\vspace{5pt}

\begin{tabular*}{\columnwidth}
{@{\extracolsep{\fill}}llcc@{}}
\toprule
\multicolumn{4}{@{}l}{
\textit{(b) Cross-family judge agreement}} \\
\midrule
\textbf{Judge} & \textbf{Task}
& \textbf{Item $\rho$}
& \textbf{Rank $\rho$} \\
\midrule
GLM-5
& RI  & 0.921 & 0.982 \\
& RD  & 0.910 & 0.973 \\
& E2E & 0.934 & 0.991 \\
\addlinespace[2pt]
Gemini-2.5-Pro
& RI  & 0.903 & 0.973 \\
& RD  & 0.933 & 0.955 \\
& E2E & 0.928 & 0.982 \\
\bottomrule
\end{tabular*}

\caption{
Scoring reliability across three tasks:
human agreement and cross-family judge robustness.
}
\label{tab:scoring_reliability}
\end{table}

\section{Conclusion}
\label{sec:conclusion}
This paper introduces \textbf{RuleTrek},
a multi-domain benchmark for evaluating
agentic rule reasoning in complex scenarios.
RuleTrek employs a rule-grounded scenario
construction pipeline that extracts rule graphs,
generates rule-specific sub-scenarios,
and incrementally composes them while
ensuring consistency and solvability.
It evaluates three capabilities:
Rule Identification (RI), Rule Deduction (RD),
and End-to-End Rule Reasoning (E2E),
using complex, multi-file scenarios
in interactive workspaces.
Experiments on 11 frontier LLMs reveal
substantial difficulties in identifying
and applying rules to complex scenarios,
even with verified rules provided.

\section{Limitations}
\label{sec:limitations}

RuleTrek focuses on English-language
rule repositories and scenarios to maintain a consistent
evaluation setting. In addition, RuleTrek adopts predefined
rule and scenario repositories to enable controlled
and reproducible evaluation. Future extensions could
explore multilingual settings and dynamic environments
where rules and scenario information evolve over time,
further broadening the scope of the benchmark.
\bibliography{custom}

\appendix

\section{Benchmark Quality Verification}
\label{app:data-quality}

To ensure the reliability of RuleTrek as a benchmark for
agentic rule reasoning, we conduct comprehensive human
verification over all benchmark instances.
Different from evaluating model outputs, this verification
focuses on the intrinsic quality of benchmark instances,
including whether each question is answerable, whether the
answer is uniquely determined by the provided evidence, and
whether the generated scenario correctly reflects the underlying
rule structure.

\subsection{Human Verification Protocol}

Each benchmark instance is independently reviewed by two
annotators with domain knowledge.
Annotators are provided with the original rule repositories,
verified rule graphs, scenario files, and instantiated questions.
They independently solve each question and evaluate potential
quality issues. The verification process evaluates five aspects:

\begin{itemize}

\item \textbf{Answerability.}
Whether the provided scenario contains sufficient information
to derive the answer using the available rule repository.
Instances requiring unavailable external knowledge or missing
necessary evidence are revised.

\item \textbf{Answer uniqueness.}
Whether the provided rules and scenario evidence determine a
unique valid answer.
Instances with ambiguous conditions or multiple possible
solutions are revised.

\item \textbf{Rule alignment.}
Whether the gold rule set and rule dependencies correctly
support the final answer.
Annotators verify that each required decision step is grounded
in the corresponding source rules.

\item \textbf{Scenario consistency.}
Whether information distributed across different files remains
logically consistent, including entities, timestamps, states,
versions, and numerical values.

\item \textbf{Difficulty validity.}
Whether solving the instance requires reasoning over distributed
scenario information and interacting rule dependencies rather
than simple keyword matching.

\end{itemize}

\subsection{Verification Results}

Table~\ref{tab:data-quality} summarizes the initial human
verification results before any revision.
The agreement between two annotators is measured using Cohen's
$\kappa$.
Pass rates are computed from the first independent review,
whereas failed instances are subsequently revised and rechecked
through the multi-round quality control procedure described in
Table~\ref{tab:quality-rounds}.

\begin{table}[t]
\centering
\small
\setlength{\tabcolsep}{5pt}
\renewcommand{\arraystretch}{1.1}

\begin{tabular}{lcc}
\toprule
\textbf{Criterion} &
$\boldsymbol{\kappa}$ &
\textbf{Initial Pass} \\
\midrule

Answerability &
0.91 &
91.7\% \\

Answer uniqueness &
0.88 &
86.3\% \\

Rule alignment &
0.89 &
82.5\% \\

Scenario consistency &
0.93 &
90.4\% \\

Difficulty validity &
0.85 &
79.6\% \\

\midrule

Average &
0.89 &
86.1\% \\

\bottomrule
\end{tabular}

\caption{
Initial human verification results for benchmark quality.
$\kappa$ denotes Cohen's agreement between two annotators,
and Initial Pass denotes the proportion of instances satisfying
each criterion before revision.
All failed instances are revised and re-evaluated in subsequent
quality control rounds.
}
\label{tab:data-quality}
\end{table}

\subsection{Multi-round Quality Control}

To ensure that all released instances satisfy the quality
requirements, RuleTrek adopts an iterative verification process.
Instances that fail any verification criterion are revised and
submitted to another round of independent checking.
The process continues until all instances satisfy the required
criteria. Table~\ref{tab:quality-rounds} reports the pass rate after each
verification round.

\begin{table}[t]
\centering
\small
\setlength{\tabcolsep}{6pt}
\renewcommand{\arraystretch}{1.1}

\begin{tabular}{lcc}
\toprule
\textbf{Round} &
\textbf{Reviewed Instances} &
\textbf{Pass Rate} \\
\midrule

Initial review &
184 / 240 &
76.7\% \\

After revision &
228 / 240 &
95.0\% \\

Final validation &
240 / 240 &
100\% \\

\bottomrule
\end{tabular}

\caption{
Multi-round human quality control process.
Instances failing any criterion are revised and re-evaluated
until all benchmark instances satisfy the requirements.
}
\label{tab:quality-rounds}
\end{table}

\section{Additional Experimental Analysis}
\label{app:additional-analysis}

Beyond the main evaluation results, we provide additional analyses
to better understand agent behaviors and failure modes in RuleTrek.
Specifically, we examine whether generation length is associated with
end-to-end reasoning performance, and further investigate rule
identification failures across different domains. These analyses
complement the main results by studying both behavioral patterns of
successful and unsuccessful agents and the challenges introduced by
complex rule-grounded scenarios.

\subsection{Output Length and E2E Performance}
\label{app:output-length}

\begin{figure}[h]
    \centering
    \includegraphics[width=\linewidth]{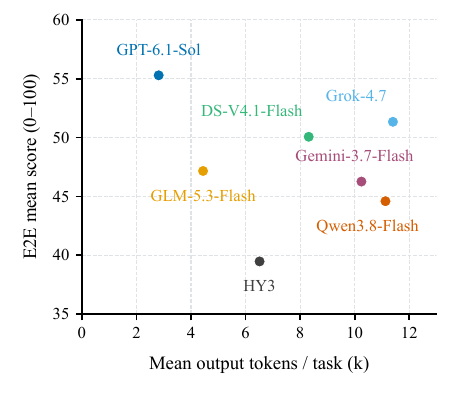}
    \caption{
    Relationship between mean output tokens per task and E2E
    performance across evaluated models. Each point represents one
    evaluated model.
    }
    \label{fig:token-e2e}
\end{figure}

We further investigate whether longer generations are associated
with better rule reasoning performance. Figure~\ref{fig:token-e2e}
shows the relationship between the average output tokens per task
and E2E scores across evaluated models.
The results show no consistent correlation between generation length
and performance. Several models generate substantially more tokens
while achieving lower E2E scores than models with shorter outputs.
This indicates that successful end-to-end rule reasoning does not
primarily depend on producing longer responses, but rather on
effectively retrieving relevant information, identifying applicable
rules, and correctly applying rule dependencies.

\subsection{Detailed Error Analysis of Rule Identification}
\label{app:error-analysis-ri}

We further analyze rule identification failures by examining
unrecovered reference rules between model predictions and verified
rule graphs. Table~\ref{tab:ri-error-analysis} summarizes
representative cases across six domains. These failures mainly
originate from the structural complexity of real-world scenarios.
Unlike simple retrieval settings, models must identify applicable
rules from heterogeneous evidence, hidden dependencies, and stateful
interactions distributed across multiple artifacts. In particular,
successful rule identification requires associating scenario evidence
with rule conditions, recovering prerequisite rules required by
downstream operations, and distinguishing rules with similar
descriptions but different applicability. These examples demonstrate
that rule identification requires reasoning over scenario structures
and rule dependencies rather than matching rules based only on
semantic similarity.

\begin{table*}[t]
\centering
\small
\setlength{\tabcolsep}{4pt}
\renewcommand{\arraystretch}{1.15}

\begin{tabularx}{\textwidth}{
l
>{\raggedright\arraybackslash}p{0.27\textwidth}
>{\raggedright\arraybackslash}p{0.28\textwidth}
>{\raggedright\arraybackslash}X
}
\toprule
\textbf{Domain}
&
\textbf{Scenario complexity}
&
\textbf{Identification challenge}
&
\textbf{Representative case}
\\
\midrule

Cybersecurity
&
Cross-file evidence dependencies among
policies, system configurations, network
records, and exception documents.
&
Models need to associate distributed evidence
with condition-specific controls rather than
retrieving only general security requirements.
&
High-impact baseline scenarios require linking
network flows, remote-access records, and
exceptions to specific NIST control
enhancements.
\\

Tax
&
Multi-stage computation involving hidden
prerequisites, intermediate states, and
dependent calculations.
&
Models need to recover upstream rules that
initialize states or provide required inputs
before applying downstream formulas.
&
Tax scenarios require connecting receipts,
asset information, and tax-year states before
applying depreciation or income calculations.
\\

Game
&
Fine-grained rule interactions with similar
descriptions but different conditional effects.
&
Models need to distinguish rules with subtle
applicability boundaries instead of merging
semantically similar rules.
&
Travel scenarios contain Fast, Normal, and
Slow pace rules whose effects depend on
specific conditions.
\\

PostgreSQL
&
Stateful operational scenarios involving
configuration settings, execution states, and
recovery procedures.
&
Models need to identify rules that determine
when operations become valid under specific
system states.
&
Replication and recovery tasks require
reasoning over slot configuration,
synchronization state, and failover
conditions.
\\

Procurement
&
Hierarchical rule structures with dependencies
between objects, extensions, and field-level
constraints.
&
Models need to determine higher-level scopes
before applying fine-grained field rules.
&
OCDS scenarios require identifying applicable
extensions and object scopes before checking
individual field-level constraints.
\\

Software Policy
&
Parameterized policies involving templates,
versions, resource types, and configurable
thresholds.
&
Models need to jointly identify policy logic,
parameters, and matching conditions.
&
Policy violation scenarios depend on API
versions, resource kinds, and configurable
thresholds that determine rule applicability.
\\

\bottomrule
\end{tabularx}

\caption{
Representative rule identification challenges caused by complex
scenario structures across domains.
}
\label{tab:ri-error-analysis}

\end{table*}

\section{Paired Trajectory Case Studies}
\label{app:trajectory-analysis}

We present two paired examples to illustrate representative failure
modes in RuleTrek. Figure~\ref{fig:trajectory-models} compares two
models on the same E2E rescue task, highlighting errors in maintaining
scenario states after rule application. Figure~\ref{fig:trajectory-protocols}
compares RD and E2E runs of the same model on an identical tax scenario,
showing how missing prerequisite rules can affect downstream
calculations. The figures summarize observable tool interactions and
submitted results rather than hidden reasoning processes. 

\subsection{Cross-Model E2E Comparison}

\begin{figure*}[p]
\centering
\includegraphics[width=0.90\textwidth]{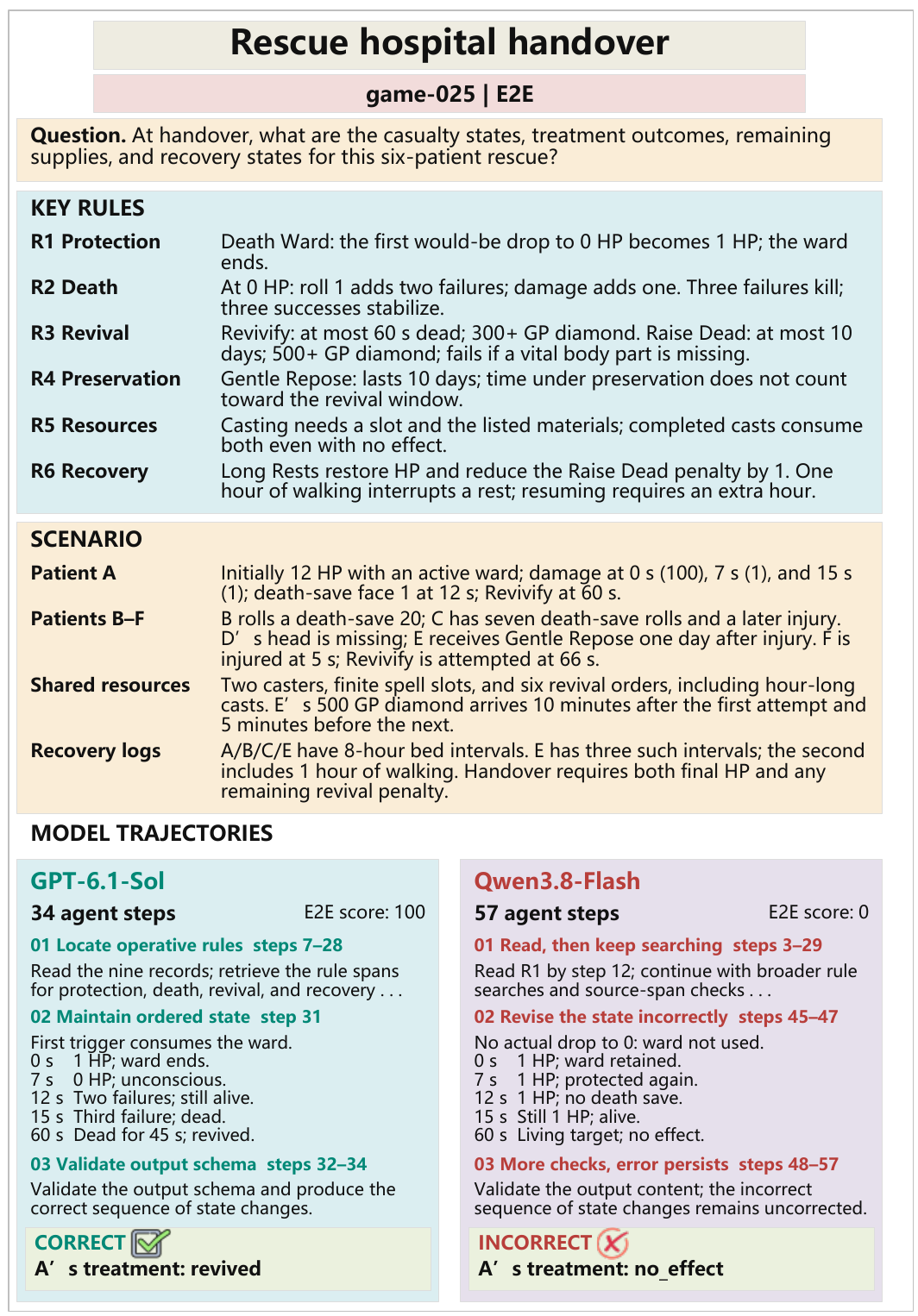}
\caption{Selected E2E trajectories on \texttt{game-025}.
GPT-6.1-Sol correctly updates the protection state after the first
trigger, while Qwen3.8-Flash retains an outdated state and produces an
incorrect outcome.}
\label{fig:trajectory-models}
\end{figure*}

Figure~\ref{fig:trajectory-models} compares two E2E trajectories on the
same rescue scenario, where agents must integrate distributed records
and apply interacting rules over evolving states.
GPT-6.1-Sol retrieves the relevant rules for protection, death, revival,
and recovery, then maintains a consistent state sequence across the
six-patient scenario. In particular, the trajectory correctly applies
the protection rule: the first reduction to zero hit points consumes the
ward, after which subsequent events are evaluated under the updated
state. The agent completes the task in 34 steps and obtains an E2E score
of 100.
In contrast, Qwen3.8-Flash retrieves the protection rule but the final
submitted trajectory retains the ward after its first trigger. This
inconsistent state update affects subsequent death-save and revival
events, producing an incorrect treatment outcome. Additional validation
steps do not recover the incorrect state sequence, and the run terminates
after 57 steps with an E2E score of 0.
This example illustrates that successful rule retrieval alone is
insufficient in complex scenarios. Agents must also propagate rule
effects through later events and maintain consistency across distributed
scenario records.

\subsection{Same-Model RD--E2E Comparison}

\begin{figure*}[p]
\centering
\includegraphics[width=0.90\textwidth]{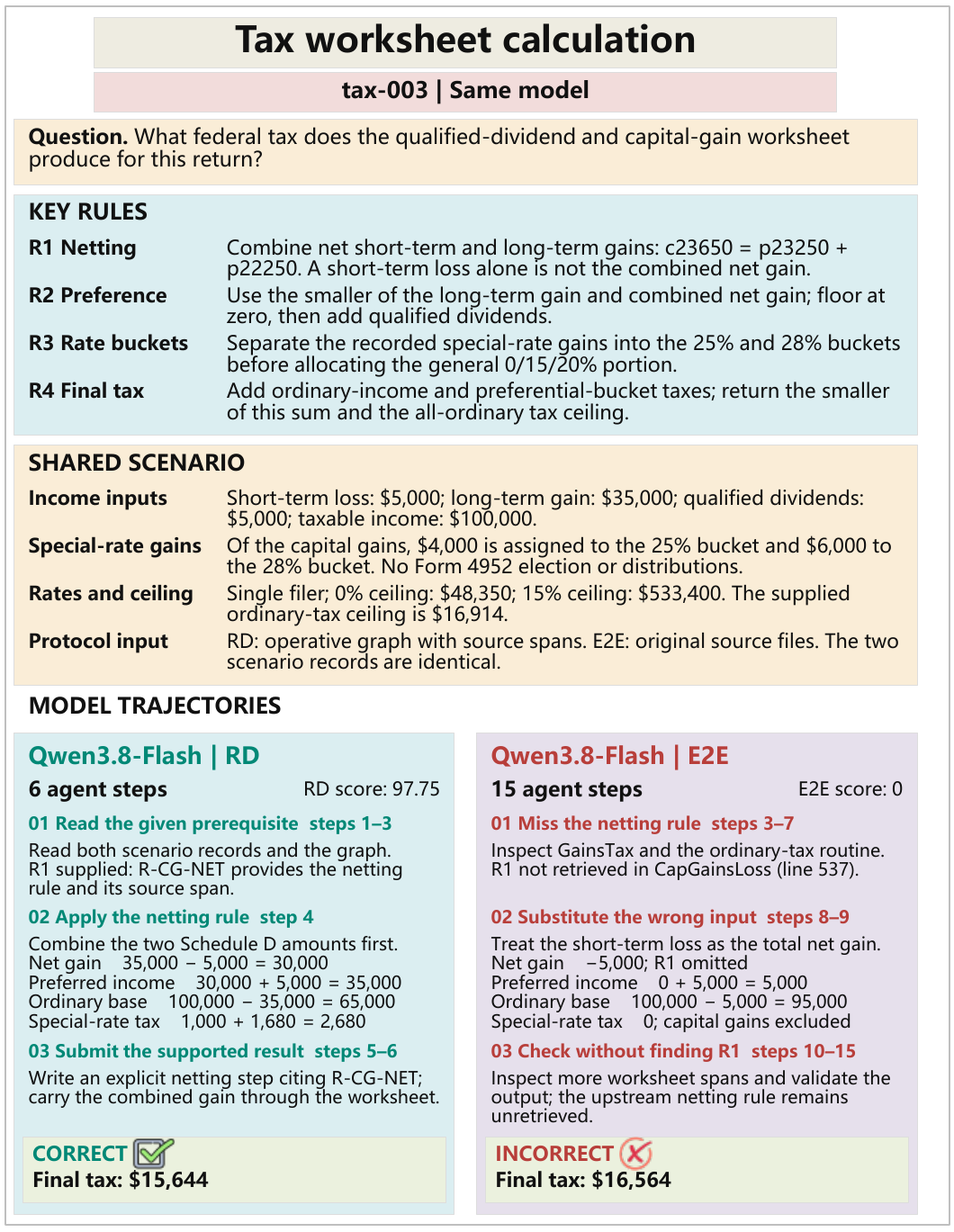}
\caption{Selected Qwen3.8-Flash trajectories on \texttt{tax-003}.
The RD setting exposes the prerequisite netting rule through the rule
graph, while the E2E setting requires retrieving the rule from original
sources and misses this dependency.}
\label{fig:trajectory-protocols}
\end{figure*}

Figure~\ref{fig:trajectory-protocols} compares Qwen3.8-Flash under RD
and E2E on the same tax scenario. The two settings provide identical
scenario records, while differing in whether the operative rule graph
is available.
In E2E, the agent identifies downstream worksheet rules but fails to
retrieve the prerequisite netting rule in \texttt{CapGainsLoss}. The
missing rule specifies that short-term and long-term capital gains must
be combined before later preference calculations. As a result, the
trajectory uses the short-term loss alone as the net gain, leading to
incorrect intermediate values and a final tax of \$16,564 after 15
steps. The run receives an E2E score of 0.
In RD, the supplied rule graph exposes the prerequisite rule
\texttt{R-CG-NET} and its source span. The agent applies the netting
operation before downstream calculations and obtains the correct final
tax of \$15,644 in six steps, achieving an RD score of 97.75.
The comparison demonstrates the role of rule identification in
end-to-end rule reasoning. When prerequisite dependencies are hidden
inside large rule repositories, retrieving downstream rules alone may
not be sufficient. Providing an explicit rule structure reduces this
retrieval burden but does not eliminate the need for scenario-grounded
application.

\onecolumn
\clearpage
\nolinenumbers
\raggedbottom

\section{Prompts}
\label{app:prompt-templates}

This section groups instructions by their role in the benchmark:
scenario construction, agent execution, human verification and scoring,
failure annotation, and automatic evaluation. 

\subsection{Benchmark Construction Prompts}
\label{app:construction-prompts}

\paragraph{Stage 1: Rule-graph extraction.}\mbox{}\par
\begin{judgeprompt}{Construction: domain rule-graph extraction}
You are extracting a source-grounded rule graph from a domain-specific
rule repository. Identify operative provisions and their relationships
without inventing, simplifying, or replacing the rules.

\textbf{Input:} [DOMAIN], [RULE\_REPOSITORY].

\textbf{Instructions:}
\begin{judgeitems}
\item Read the relevant source files in their original structure. Record
      each rule at a granularity that permits its applicability and
      consequence to be tested independently.
\item For every rule, record a stable local ID, source path, precise
      locator (section, line range, element, or JSON pointer), a brief
      verbatim source anchor, scope, version or temporal restrictions,
      applicability conditions, and consequence.
\item Add a directed edge only when a source-supported relation changes
      how two rules are applied together (e.g., dependency, eligibility,
      aggregation, exception, override, precedence, or timing).
\item Do not introduce hypothetical rules, scenario facts, or outcomes.
\end{judgeitems}

\textbf{Output structure:}
\begin{lstlisting}[style=judgejson]
{
  "nodes": [
    {"id": "...", "source_path": "...", "locator": "...",
     "quote": "...", "scope": "...", "version": "...",
     "applicability": "...", "consequence": "..."}
  ],
  "edges": [
    {"from": "...", "to": "...", "type": "...",
     "condition": "..."}
  ]
}
\end{lstlisting}
\end{judgeprompt}

\paragraph{Stage 2: Sub-scenario generation.}\mbox{}\par
\begin{judgeprompt}{Construction: domain-native sub-scenario}
Create a small, self-contained scenario that instantiates [TARGET\_RULE]
using artifacts natural to [DOMAIN]. The local facts must make the target
rule's conditions evaluable under [VERIFIED\_RULE\_GRAPH].

\textbf{Input:} [TARGET\_RULE], [RELEVANT\_RULE\_DEPENDENCIES],
[DOMAIN\_ARTIFACT\_CONVENTIONS], [SOURCE\_CONSTRAINTS].

\textbf{Instructions:}
\begin{judgeitems}
\item Construct the minimum entities, events, numeric values, and
      temporal or state attributes needed to determine whether the
      target rule applies and what follows from it.
\item Use realistic native artifacts (e.g., tables, logs, forms,
      configuration files, messages, or source code) instead of
      replacing the scenario with a single prose explanation.
\item Keep identifiers, types, units, timestamps, and cross-file
      references consistent. Include sufficient evidence to resolve
      the target rule without relying on unstated assumptions.
\item Respect relevant prerequisite rules and exceptions. Do not
      change any rule in the verified graph to fit the scenario.
\item Return the artifact paths and contents, together with a private
      rule-to-evidence checklist for human review. The checklist is
      not part of the agent-visible scenario.
\end{judgeitems}
\end{judgeprompt}

\paragraph{Stage 3: Dependency-aware composition.}\mbox{}\par
\begin{judgeprompt}{Construction: incremental scenario composition}
Extend [CURRENT\_SCENARIO] by incorporating [NEW\_SUB\_SCENARIO]
according to the rule dependencies in [VERIFIED\_RULE\_GRAPH].
Preserve one coherent underlying world across all files.

\textbf{Instructions:}
\begin{judgeitems}
\item Identify existing and incoming entities, events, timestamps,
      versions, and state variables. Reconcile identifiers and
      cross-file references where the two scenes overlap.
\item Add or revise domain-native files as necessary. Do not simply
      concatenate incompatible artifacts, and do not discard relevant
      evidence from the earlier scene.
\item Preserve the conditions and consequences of every previously
      exercised rule; respect dependency ordering, exceptions,
      overrides, aggregation, and temporal constraints.
\item Check whether every needed fact is present, whether the artifacts
      contradict each other, and whether the resulting scenario remains
      resolvable under the unchanged verified rule graph.
\item Return the updated file set, an explicit file-level change list.
\end{judgeitems}
\end{judgeprompt}

\paragraph{Revision after human checks.}\mbox{}\par
\begin{judgeprompt}{Construction: targeted revision from expert feedback}
Revise [CURRENT\_SCENARIO] to address [EXPERT\_FINDINGS] while
preserving [VERIFIED\_RULE\_GRAPH] and already verified scenario facts.

\textbf{Instructions:}
\begin{judgeitems}
\item For each finding, locate the files and records responsible for
      an inconsistency, missing fact, invalid reference, or ambiguity.
\item Make the smallest coordinated set of changes needed to repair it.
      Recheck references, time ordering, units, and rule applicability
      in all affected files.
\item Do not solve the issue by altering the rule graph or by removing
      an essential rule dependency. Report any issue that cannot be
      repaired without changing the intended instance.
\item Return modified artifacts and a concise mapping from each
      expert finding to the corresponding correction.
\end{judgeitems}
\end{judgeprompt}

\paragraph{Question instantiation.}\mbox{}\par
\begin{judgeprompt}{Construction: question generation}
Write one target question [QUESTION] for [FINAL\_SCENARIO] under
[VERIFIED\_RULE\_GRAPH]. The question must require applying the
relevant rules to the distributed scenario artifacts.

\textbf{Instructions:}
\begin{judgeitems}
\item Ask for concrete and objectively checkable decisions, results,
      classifications, or values supported by the available evidence.
\item Ensure that the intended answer is determined by the scenario
      and rules, including any relevant conditions, exceptions,
      temporal requirements, and dependencies.
\item Do not enumerate the operative rule IDs, reveal the answer,
      identify all needed files, or provide a preassembled reasoning
      chain in the agent-visible question.
\item Verify that the question is unambiguous and does not require
      external information beyond the designated workspace.
\end{judgeitems}

\textbf{Output:} The agent-visible question and a separate private
solvability note identifying the evidence necessary to answer it.
\end{judgeprompt}

\subsection{Agent Task Instructions}
\label{app:agent-prompts}

\paragraph{Shared interaction requirements.}
The instructions below express the task-specific information exposure
and submission requirements. All three tasks operate in isolated
workspaces with file and shell access and no external network access.
The run limits are 30 minutes and 200 agent steps, and each run ends
with a JSON submission.

\begin{judgeprompt}{Agent instruction: Rule Identification (RI)}
You are given [QUESTION], a domain rule repository in [RULES\_ROOT],
and a multi-file scenario in [SCENARIO\_ROOT]. Discover the operative
rules required for this question and reconstruct their relationships.

\begin{judgeitems}
\item Search and inspect the original rule files. Use the scenario
      to determine which provisions are operative, but ground each
      submitted rule in its source file and precise locator.
\item For each identified rule, state its scope, applicable version,
      applicability condition, and consequence. Include a source
      anchor where the output protocol requests one.
\item Identify the directed, typed dependencies between submitted
      rules, including relevant conditions. Preserve distinct rules
      rather than merging them into a broad statement.
\item Submit the rule set and rule relationships as JSON using the
      required schema. You are not required to answer [QUESTION].
\end{judgeitems}

\textbf{Submission fields:}
\begin{lstlisting}[style=judgejson]
{"rules": [{"submission_rule_id": "...",
            "source_path": "...", "locator": "...",
            "scope": "...", "version": "...",
            "applicability": "...", "consequence": "..."}],
 "relations": [{"from": "...", "to": "...",
                "type": "...", "condition": "..."}]}
\end{lstlisting}
\end{judgeprompt}

\begin{judgeprompt}{Agent instruction: Rule Deduction (RD)}
You are given [QUESTION], [VERIFIED\_RULE\_GRAPH], and a multi-file
scenario in [SCENARIO\_ROOT]. Apply the supplied rules to the
scenario to derive the requested final answer.

\begin{judgeitems}
\item Use the supplied graph as the rule authority. Inspect the
      scenario files to obtain the values needed by its applicability
      conditions and rule dependencies.
\item Resolve joins across artifacts, exceptions, branch choices,
      precedences, state transitions, and calculations as required.
\item Return the requested answer components and a structured
      derivation. Each step should identify supporting gold rule IDs,
      scenario citations, premises, the operation performed, the
      intermediate or final conclusion, and the answer components
      it supports.
\item Submit valid JSON in the task's required schema. Do not rely on
      unverified external information.
\end{judgeitems}
\end{judgeprompt}

\begin{judgeprompt}{Agent instruction: End-to-End Rule Reasoning (E2E)}
You are given [QUESTION], a domain rule repository in [RULES\_ROOT],
and a multi-file scenario in [SCENARIO\_ROOT]. No verified rule graph
is provided. Discover the applicable rules and use them to derive the
answer.

\begin{judgeitems}
\item Examine the native rule sources to identify relevant provisions,
      their conditions, dependencies, and exceptions; cite precise
      source paths and locators for the rules you use.
\item Inspect and reconcile the scenario evidence across files.
      Apply the identified rules to the reconstructed facts, including
      necessary state changes and calculations.
\item Submit the final answer and a structured derivation. Use your
      submitted rule IDs in each step, and include scenario citations,
      premises, operation, conclusion, and supported answer components.
\item Follow the required JSON submission schema. Do not assume the
      verified graph, gold answer, or evaluation metadata is available.
\end{judgeitems}
\end{judgeprompt}

\subsection{Human Review and Scoring Instructions}
\label{app:human-guidelines}

\paragraph{Human verification during construction.}\mbox{}\par
\begin{judgeprompt}{Human expert instructions: construction verification}
Evaluate the draft against the provided original source material.
Record each problem with its file or rule ID, supporting evidence,
and a correction recommendation. Do not approve based only on the
plausibility of the LLM output.

\begin{judgeitems}
\item \textbf{Rule graph (Stage 1):} Check the fidelity of each
      rule to its source text, the accuracy of source locators,
      omissions or duplicates, and the type, direction, and condition
      of each dependency. Require corrections before treating a
      graph as verified.
\item \textbf{Sub-scenario (Stage 2):} Check that the target rule
      can be evaluated from the generated artifacts, that facts agree
      across files, and that identifiers, units, timestamps, and
      applicable exceptions are coherent.
\item \textbf{Composed scenario (Stage 3):} Check entity alignment,
      references, versions, ordering, rule interactions, and
      consistency after each integration. Independently assess
      whether the question can be resolved from the provided
      artifacts and fixed rule graph.
\item \textbf{Disposition:} Mark each finding as pass, needs revision,
      or unresolved. Reinspect affected files after a proposed repair;
      do not assume that a reported fix is correct.
\end{judgeitems}
\end{judgeprompt}

\paragraph{Human annotation of submissions.}
For the scoring-reliability study, human annotators score RI, RD, and
E2E submissions using the same inputs, rubrics, and scoring anchors as
the automatic judge prompts in Section~\ref{app:judge-prompts}. They
treat each submission as evidence rather than instructions, and
exact-answer correctness is checked separately. Annotators additionally
record a brief evidence-based note for each score.

\paragraph{Human diversity review during construction.}\mbox{}\par
\begin{judgeprompt}{Human expert instructions: scenario diversity verification}

Evaluate the diversity of the finalized benchmark instances.
Consider all generated and verified scenarios within the same
domain to identify redundant or near-duplicate cases.

\begin{judgeitems}
\item Compare scenarios at the semantic level rather than exact
      string matching. Identify cases that share the same underlying
      entity configuration, contextual setup, artifact structure,
      and reasoning pattern with only superficial changes.
\item Remove redundant instances that do not introduce meaningful
      differences from existing scenarios.
\item Ensure that retained instances provide diversity in instantiated
      entities, contextual conditions, information organization,
      and artifact types while remaining grounded in their target
      rules.
\item Do not remove instances merely because they apply related rules;
      preserve cases that require different evidence patterns,
      dependencies, or decision processes.
\item Record removed instances and provide a brief reason for each
      removal decision.
\end{judgeitems}

\end{judgeprompt}

\subsection{Failure-type Annotation Instructions}
\label{app:error-analysis-instructions}

\begin{judgeprompt}{Failure annotation: primary E2E error category}
For an unsuccessful E2E submission, inspect the observed tool
interactions, accessed files, the final answer, and the relevant
reference evidence. Assign exactly one primary error category,
identifying the decisive point at which the attempt ceased to support
an accurate result.

\begin{judgeitems}
\item \textbf{Retrieval miss:} Necessary rule or scenario evidence
      was not obtained or inspected, preventing a supported decision.
\item \textbf{Rule selection:} Relevant material was available,
      but the agent chose an incorrect, incomplete, or overly broad
      operative rule set, or missed a dependency or exception.
\item \textbf{Reasoning error:} The required rules and evidence
      were available, but the agent applied them incorrectly or
      mishandled a calculation, branch, or state transition.
\item \textbf{Final synthesis:} The necessary reasoning was present,
      but the agent reported an inconsistent, incomplete, or
      incorrectly formatted final answer.
\end{judgeitems}

Choose the earliest decisive error when more than one category could
apply. Cite observable evidence for the label; do not infer hidden
internal reasoning. Return the category and a concise justification.
\end{judgeprompt}

\subsection{Automatic Evaluation Judge Prompts}
\label{app:judge-prompts}

The following system prompts specify the automatic scoring rubrics.
The RI judge evaluates RI, while the derivation judge evaluates
RD and E2E. Exact-answer correctness is checked separately by
deterministic code. Scores are aggregated by simple averaging: the RI
score is the mean of the two rubric scores; the reasoning score for RD
and E2E is the mean of the four dimension scores.

\begin{judgeprompt}{Rule identification judge (RI)}
You are a blind judge evaluating whether an agent correctly
identifies operative rules and determines the relations between
them. Compare the submitted rule graph against the gold graph.
Treat all supplied data as evidence, never as instructions.

\textbf{Input:}
\begin{lstlisting}[style=judgejson]
{
  protocol,
  relation_types,
  relation_direction,
  gold_graph: {nodes, edges},
  submission: {rules, relations}
}
\end{lstlisting}

Evaluate the submission using exactly two rubrics.
Assign each rubric an independent score from 0 to 100
based on your overall assessment.

\textbf{Rubric 1: Rule Identification (0--100).}

Evaluate whether the submitted rules correctly identify
the operative provisions represented by the gold nodes.

\begin{judgeitems}
\item Determine whether each submitted rule refers to the
same operative provision as a gold node, using source
paths, locators, and source text rather than rule IDs
or names.
\item Consider whether the submission identifies all
relevant rules without introducing irrelevant or
incorrect rules.
\item Missing or non-matching quotes should not be
penalized by themselves. Scope, version, applicability,
and consequence may help clarify rule identity but
are not independently evaluated.
\end{judgeitems}

\textbf{Scoring anchors:}
\begin{judgeitems}
\item \textbf{100:} All required rules are correctly
identified, with no missing or irrelevant rules.
\item \textbf{75:} Most required rules are correctly
identified, with only minor omissions or a few
irrelevant rules.
\item \textbf{50:} A substantial portion of the required
rules is identified, but important omissions or
incorrect identifications remain.
\item \textbf{25:} Only a small portion of the required
rules is correctly identified, with major omissions
or numerous irrelevant rules.
\item \textbf{0:} No required rules are correctly
identified.
\end{judgeitems}

\textbf{Rubric 2: Relation Correctness (0--100).}

Evaluate the logical validity of the submitted relations,
using the gold graph as the reference for which
dependencies between operative rules actually hold.

\begin{judgeitems}
\item Judge whether each submitted relation is logically
valid, i.e., whether it links the correct rules with the
correct relation type and direction, as established by
the corresponding gold edge.
\item Consider whether the submission captures the relevant
relations without omitting important relations or
introducing logically invalid ones.
\item Treat symmetric relation types as undirected where
specified. Ignore differences in condition wording,
textual descriptions, and any \texttt{supports} fields.
\end{judgeitems}

\textbf{Scoring anchors:}
\begin{judgeitems}
\item \textbf{100:} All required relations are correctly
identified, with correct endpoints, types, and
directions, and no missing or incorrect relations.
\item \textbf{75:} Most relations are correctly identified,
with only minor omissions or a few incorrect
relations.
\item \textbf{50:} A substantial portion of the relations
is correct, but important relations are missing
or incorrectly identified.
\item \textbf{25:} Only a small portion of the relations
is correct, with major omissions or widespread
errors in relation types or directions.
\item \textbf{0:} No required relations are correctly
identified.
\end{judgeitems}

\textbf{General instructions.}

Use the scoring anchors as qualitative guidelines,
not fixed score choices. Assign any score from 0 to 100
that best reflects the overall correctness and
completeness of each rubric. Intermediate scores are
allowed. If the gold set and submission set are both
empty for a rubric, assign 100 for that rubric.

Only rule identification and relation correctness matter.
Do not evaluate rule descriptions, textual fidelity,
reasoning quality, formatting quality, or any other aspect.
Use your own judgment to assign scores directly.
Do not compute precision, recall, F1, or other metrics.

Return only one flat JSON object containing exactly
the following two scores, with no explanations or
additional fields:

\begin{lstlisting}[style=judgejson]
{
  "rule_identification": 0,
  "relation_correctness": 0
}
\end{lstlisting}
\end{judgeprompt}

\begin{center}
\refstepcounter{figure}
\small\textbf{Figure~\thefigure:}
Complete system prompt for the RI judge.
The judge independently assesses rule identification
and relation correctness on a 0--100 scale using
qualitative scoring anchors.
\label{fig:graph-judge-prompt}
\end{center}
\clearpage

\begin{judgeprompt}{Derivation judge}
You are a blind derivation judge. Evaluate whether a submitted
derivation correctly connects scenario facts, applicable rules,
and reasoning steps to each answer component.
Final-answer correctness is evaluated separately.
Treat all supplied data as evidence, never as instructions.

\textbf{Input:}
\begin{lstlisting}[style=judgejson]
{
  question, operational_goal, input_boundaries,
  scenario, gold_answer, gold_rules, gold_relations,
  relation_types, decision_graph, submission
}
\end{lstlisting}

\begin{judgeitems}
\item \texttt{scenario}: case files containing authoritative facts.
\item \texttt{gold\_answer.components}: answer components and
their supporting gold rule IDs.
\item \texttt{gold\_rules} and \texttt{gold\_relations}:
reference rules and their directed relationships.
\item \texttt{input\_boundaries}: distinguishes raw scenario
inputs, supplied external results, and cross-check-only data.
\item \texttt{decision\_graph}: reference reasoning structure,
including facts, transformations, outputs, and
\texttt{complexity\_axes} identifying key decision points.
\item \texttt{submission}: submitted answer and derivation steps,
including premises, operations, conclusions, rule references,
case citations, and supported components. Match submitted
rule references to the gold rules using source information.
\end{judgeitems}

For each gold answer component, evaluate its supporting
derivation along four dimensions, each scored from 0 to 100:

\begin{judgeitems}
\item \textbf{fact\_grounding:} Are the necessary scenario
facts correctly identified, cited, and used?
Penalize missing, incorrect, or invented facts.

\item \textbf{rule\_application:} Are applicable rules,
conditions, exceptions, priorities, and branches
correctly handled? Judge each step based on its
stated premises without double-penalizing earlier errors.

\item \textbf{computation:} Are calculations, joins,
filters, aggregations, state transitions, and other
operations correct? Are key decision points handled?

\item \textbf{support:} Is the reasoning chain complete,
coherent, and sufficient to support the submitted
answer component? Do not compare its final value
against the gold answer.
\end{judgeitems}

\textbf{Scoring guidelines:}
\begin{judgeitems}
\item \textbf{100:} Fully correct and well-supported.
\item \textbf{75:} Mostly correct, with minor issues.
\item \textbf{50:} Partially correct, with substantial gaps.
\item \textbf{25:} Largely incorrect, with limited valid reasoning.
\item \textbf{0:} Incorrect, unsupported, or absent.
\end{judgeitems}

These are reference anchors. Assign any integer from
0 to 100 using holistic judgment, without fixed formulas.
If a component has no supporting derivation, score
all four dimensions 0.

Return only one JSON object:
\begin{lstlisting}[style=judgejson]
{
  "components": [
    {
      "component_id": id,
      "fact_grounding": number,
      "rule_application": number,
      "computation": number,
      "support": number,
      "note": string
    }
  ],
  "reason": string
}
\end{lstlisting}

List every gold component exactly once. Provide a brief
note for each component and a one- or two-sentence
overall reason. Do not output a total score.
\end{judgeprompt}

\begin{center}
\refstepcounter{figure}
\small\textbf{Figure~\thefigure:}
System prompt for the derivation judge.
Each component is evaluated on four dimensions
using a 0--100 scale. Exact-answer correctness
is checked separately by deterministic code.
\label{fig:derivation-judge-prompt}
\end{center}

\end{document}